# REAL TIME FACE RECOGNITION USING ADABOOST IMPROVED FAST PCA ALGORITHM

K. Susheel Kumar[1], Shitala Prasad[2], Vijay Bhaskar Semwal[3], R C Tripathi[2]

[1]Department of Computer Engineering, Ideal Institute of Technology, Ghaziabad, INDIA
sus.iiita.932@gmail.com

[2]Department of Information Technology, Indian Institute of Information Technology Allahabad, INDIA.
ihc2009011@iiita.ac.in, rctripathi@iiita.ac.in

[3]PES-KM Group, Newgen Software Technology, Greater Noida, INDIA.
vijay.semwal@newgen.co.in

## ABSTRACT

*This paper presents an automated system for human face recognition in a real time background world for a large homemade dataset of persons face. The task is very difficult as the real time background subtraction in an image is still a challenge. Addition to this there is a huge variation in human face image in terms of size, pose and expression. The system proposed collapses most of this variance. To detect real time human face AdaBoost with Haar cascade is used and a simple fast PCA and LDA is used to recognize the faces detected. The matched face is then used to mark attendance in the laboratory, in our case. This biometric system is a real time attendance system based on the human face recognition with a simple and fast algorithms and gaining a high accuracy rate..*

## KEYWORDS

*Face recognition, Eigenface, AdaBoost, Haar Cascade Classifier, Principal Component Analysis (PCA), Fast PCA, Linear Discriminant Analysis (LDA).*

## 1. INTRODUCTION

Over the last ten years or so, face recognition has become a popular area of research in computer vision. Face recognition is also one of the most successful applications of image analysis and understanding. Because of the nature of the problem of face recognition, not only computer science researchers are interested in it, but neuroscientists and psychologists are also interested for the same. It is the general opinion that advances in computer vision research will provide useful insights to neuroscientists and psychologists into how human brain works, and vice versa.

The topic of real time face recognition for video and complex real-world environments has garnered tremendous attention for student to attend class daily means online attendance system as well as security system based on face recognition. Automated face recognition system is a big challenging problem and has gained much attention from last few decades. There are many approaches in this field. Many proposed algorithms are there to identify and recognize human being face form given dataset. The recent development in this field has facilitated us with fast processing capacity and high accuracy. The efforts are also going in the direction to include learning techniques in this complex computer vision technology.

There are many existing systems to identify faces and recognized them. But the systems are not so efficient to have automated face detection, identification and recognition. A lot of research work is going in this direction to increase the visual power of computer. Hence, there is a lot of scope in the development of visual and vision system. But there are difficulties in the path such as development of efficient visual feature extracting algorithms and high processing power for retrieval from a huge image database. As image is a complex high dimension (3D) matrix and processing matrix operation is not so fast and perfect. Hence, this direction us to handle with a huge image database and focus on the new algorithms which are more real-time and more efficient with maximum percentage of accuracy. Efficient and effective recognition of human face from image databases is now a requirement. Face recognition is a biometric method for identifying individuals by their features of face. Applications of face recognition are widely spreading in areas such as criminal identification, security system, image and film processing. From the sequence of image captured by the capturing device, in our case camera, the goal is to find the best match in the database. Using pre-storage database we can identify or verify one or more identities in the scene. The general block diagram for face recognition system is having three main blocks, the first is face detection, second is face extraction and the third face recognition. The basic overall face recognition model looks like the one below, in figure 1.

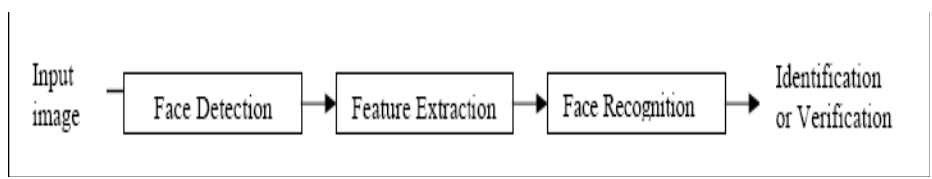

Figure 1. Basic Block Flow Diagram of Face Recognition.

Different approaches of face recognition for still images can be categorized into tree main groups such as holistic approach, feature-based approach, and hybrid approach [1]. Face recognition form a still image can have basic three categories, such as holistic approach, feature-based approach and hybrid approach [2].

1.1     Holistic Approach: - In holistic approach, the whole face region is taken as an input in face detection system to perform face recognition.

1.2     Feature-based Approach: - In feature-based approach, local features on face such as nose and eyes are segmented and then given to the face detection system to easier the task of face recognition.

1.3     Hybrid Approach: - In hybrid approach, both local features and the whole face is used as the input to the face detection system. It is more similar to the behaviour or human being to recognize the face.

This paper is divided into seven sections. The first section is the introduction part; the second section is a problem statement; the third section face recognition techniques- literature review; the fourth section is the proposed method for feature extraction form a face image dataset, the fifth division is about the implementation; the second last section shows the results; and the last is the conclusion section.

## 2. PROBLEM STATMENT

The difficulties in face recognition are very real-time and natural. The face image can have head pose problem, illumination problem, facial expression can also be a big problem. Hair style and aging problem can also reduce the accuracy of the system. There can be many other problems such as occlusion, i.e., glass, scarf, etc., that can decrease the performance. Image is a multi-dimension matrix in mathematics that can be represented by a matrix value. Image can be treated as a vector having magnitude and direction both. It is known as vector image or image vector.

If $x_i$ represents a p x q image vector and x is matrix of image vector. Thus, image matrix can be represented as x= $\{x_1, x_2, \ldots, x_n\}^t$, where t is transpose of the matrix x. Thus, to identify the glass in an image matrix is very difficult and requires some new approaches that can overcome these limitations. The algorithm proposed in this paper successfully overcomes these limitations. But before that let's see what all techniques have been used in the field of face identification and face recognition.

# 3. FACE RECOGNITION TECHNIQUES

## 3.1. Face detection

Face detection is a technology to determine the locations and size of a human being face in a digital image. It only detects facial expression and rest all in the image is treated as background and is subtracted from the image. It is a special case of object-class detection or in more general case as face localizer. Face-detection algorithms focused on the detection of frontal human faces, and also solve the multi-view face detection problem. The various techniques used to detect the face in the image are as below:

### 3.1.1. Face detection as a pattern-classification task:

In this face detection is a binary-pattern classification task. That is, the content of a given part of an image is transformed into features, after which a classifier trained on example faces decides whether that particular region of the image is a face, or not [3].

### 3.1.2. Controlled background:

In this technique the background is still or is fixed. Remove the background and only the faces will be left, assuming the image only contains a frontal face [3].

### 3.1.3. By color:

This technique is vulnerable. In this skin color is used to segment the color image to find the face in the image. But this has some drawback; the still background of the same color will also be segmented.

### 3.1.4. By motion:

The face in the image is usually in motion. Calculating the moving area will get the face segment [3]. But this too have many disadvantages as there may be backgrounds which are in motion.

### 3.1.5. Model-based:

A face model can contain the appearance, shape, and motion of faces [3]. This technique uses the face model to find the face in the image. Some of the models can be rectangle, round, square, heart, and triangle. It gives high level of accuracy if used with some other techniques.

## 3.2. Face Recognition

Face recognition is a technique to identify a person face from a still image or moving pictures with a given image database of face images. Face recognition is biometric information of a person. However, face is subject to lots of changes and is more sensitive to environmental changes. Thus, the recognition rate of the face is low than the other biometric information of a person such as fingerprint, voice, iris, ear, palm geometry, retina, etc. There are many methods for face recognition and to increase the recognition rate. Some of the basic commonly used face recognition techniques are as below:

### 3.2.1. Neural Networks

A neural network learning algorithm called Backpropagation is among the most effective approaches to machine learning when the data includes complex sensory input such as images, in our case face image. Neural network is a nonlinear network adding features to the learning system. Hence, the features extraction step may be more efficient than the linear Karhunen-Loeve methods which chose a dimensionality reducing linear projection that maximizes the scatter of all projected samples [3]. This has classification time less than 0.5 seconds, but has training time more than hour or hours. However, when the number of persons increases, the computing expense will become more demanding [5]. In general, neural network approaches encounter problems when the number of classes, i.e., individuals increases.

### 3.2.2. Geometrical Feature Matching

This technique is based on the set of geometrical features from the image of a face. The overall configuration can be described by a vector representing the position and size of the main facial features, such as eyes and eyebrows, nose, mouth, and the shape of face outline [5]. One of the pioneering works on automated face recognition by using geometrical features was done by T. Kanade [5]. Their system achieved a peak performance of 75% recognition rate on a database of 20 people using two images per person, one as the model and the other as the test image [4]. I.J. Cox el [6] introduced a mixture-distance technique which achieved 95% recognition rate on a query database of 685 individuals. In this, each of the face was represented by 30 manually extracted distances. First the matching process utilized the information presented in a topological graphics representation of the feature points. Then the second will after that will be compensating for the different center location, two cost values, that are, the topological cost, and similarity cost, were evaluated. In short, geometrical feature matching based on precisely measured distances between features may be most useful for finding possible matches in a large database [4].

### 3.2.3. Graph Matching

Graph matching is another method used to recognize face. M. Lades et al [7] presented a dynamic link structure for distortion invariant object recognition, which employed elastic graph matching to find the closest stored graph. This dynamic link is an extension of the neural networks. Face are represented as graphs, with nodes positioned at fiducial points, (i.e., exes, nose…,), and edges labeled with two dimension (2-D) distance vector. Each node contains a set of 40 complex Gabor wavelet coefficients at different scales and orientations (phase, amplitude). They are called "jets". Recognition is based on labeled graphs [8]. A jet describes a small patch of grey values in an image I (~x) around a given pixel ~x = (x; y). Each is labeled with jet and each edge is labeled with distance. Graph matching, that is, dynamic link is superior to all other recognition techniques in terms of the rotation invariance. But the matching process is complex and computationally expensive.

### 3.2.4. Eigenfaces

Eigenface is a one of the most thoroughly investigated approaches to face recognition [4]. It is also known as Karhunen-Loeve expansion, eigenpicture, eigenvector, and principal component. L. Sirovich and M. Kirby [9, 10] used principal component analysis to efficiently represent pictures of faces. Any face image could be approximately reconstructed by a small collection of weights for each face and a standared face picture, that is, eigenpicture. The weights here are the obtained by projecting the face image onto the eigenpicture. In mathematics, eigenfaces are the set of eigenvectors used in the computer vision problem of human face recognition. The principal components of the distribution of faces, or the eigenvectors of the covariance matrix of the set of face image is the eigenface. Each face can be represented exactly by a linear combination of the eigenfaces [4]. The best M eigenfaces construct an M dimension (M-D) space that is called the "face space" which is same as the image space discussed earlier.

Illumination normalization [10] is usually necessary for the eigenfaces approach. L. Zhao and Y.H. Yang [12] proposed a new method to compute the covariance matrix using three images each was taken in different lighting conditions to account for arbitrary illumination effects, if the object is Lambertian A. Pentland, B. Moghaddam [13] extended their early work on eigenface to eigenfeatures corresponding to face components, such as eyes, nose, mouth. Eigenfeatures combines facial metrics (measuring distance between facial features) with the eigenface approach [11]. This method of face recognition is not much affected by the lighting effect and results somewhat similar results in different lighting conditions.

### 3.2.5. Fisherface

Belhumeur et al [14] propose fisherfaces method by using PCA and Fisher's linear discriminant analysis to propduce subspace projection matrix that is very similar to that of the eigen space method. It is one of the most successful widely used face recognition methods. The fisherfaces approach takes advantage of within-class information; minimizing variation within each class, yet maximizing class separation, the problem with variations in the same images such as different lighting conditions can be overcome. However, Fisherface requires several training images for each face, so it cannot be applied to the face recognition applications where only one example image per person is available for training.

### 3.3. Feature Extraction Techniques

Facial feature extraction is necessary for identification of an individual face on a computer. As facial features, the shape of facial parts is automatically extracted from a frontal face image. There can be three methods for the facial feature extraction as given below:

### 3.3.1. Geometry-based

This technique is prosed by Kanada [15] the eyes, the mouth and the nose base are localized using the vertical edge map. These techniques require threshold, which, given the prevailing sensitivity, may adversely affect the achieved performance.

### 3.3.2. Template-based

This technique, matches the facial components to previously designed templates using appropriate energy functional. Genetic algorithms have been proposed for more efficient searching times in template matching.

### 3.3.3. Color segmentation techniques

This technique makes use of skin color to isolate the facial and non-facial part in the image. Any non-skin color region within the face is viewed as a candidate for eyes and or mouth.

Research and experiments on face recognition still continuing since many decades but still there is no single algorithm perfect in real time face recognition with all the limitations discussed in second section. Here, in this paper, a new approach is proposed to somewhat overcome the limitations with a very less complexity.

## 4. FACIAL FEATURE EXTRACTION

In many problem domains combining more than one technique with any other technique(s) often results in improvement of the performance. Boosting is one of such technique used to increase the performance result. Facial features are very important in face recognition. Facial features can be of different types: region [16, 17], key point (landmark) [18, 19], and contour [20, 21]. In this paper, AdaBoost: Boosting algorithm with Haar Cascade Classifier for face detection and fast PCA and PCA with LDA for the purpose of face recognition. All these algorithms are explained one by one.

## 4.1. Face Detection

### 4.1.1. AdaBoost: The Boosting Algorithm

AdaBoost is used as a short form for Adaptive Boosting, which is a widely used machine learning algorithm and is formulated by Yoav Freund and Robert Schapire. It's a meta-algorithm, algorithm of algorithm, and is used in conjunction with other learning algorithms to improve their performance of that algorithm(s) [24]. In our case abaBoost is combined with haar feature to improve the performance rate. The algorithm, AdaBoost is an adaptive algorithm in the sense that the subsequent classifiers built is tweaked in favor of instances of those misclassified by the previous classifiers. But it is very sensitive to noise data and the outliers. AdaBoost takes an input as a training set $S = \{(x_1, y_1), ..., (x_m, y_m)\}$ , where each instance of S, $x_i$, belongs to a domain or instance space $X$, and similarly each label $y_i$ belongs to the finite label space, that is $Y$. Here in this paper, we only focus on the binary case when $Y = \{-1, +1\}$ . The basic idea of boosting is actually to use the weak learner of the features calculated, to form a highly correct prediction rules by calling the weak learner repeatedly processed on the different-different distributions over the training examples.

### 4.1.2. Haar Cascade Classifier

A Haar Classifier is also a machine learning algorithmic approach for the visual object detection, originally given by Viola & Jones [23]. This technique was originally intended for the facial recognition but it can be used for any other object. The most important feature of the Haar Classifier is that, it quickly rejects regions that are highly unlikely to be contained in the object. The core basis for Haar cascade classifier object detection is the Haar-like features. These features, rather than using the intensity values of a pixel, use the change in contrast values between adjacent rectangular groups of pixels [25]. The variance of contrast between the pixel groups are used to determine relative light and dark areas. The various Haar-like-features are shown in the figure 2.a. The set of basic Haar-like-feature is shown in figure 2.b, rotating which the other features can be generated. The value of a Haar-like feature is the difference between the sum of the pixel gray level values within the black and white rectangular regions, i.e.,

$f(x) = Sum_{black\ rectangle}$ *(pixel gray level)* $- Sum_{white\ rectangle}$ *(pixel gray level)*

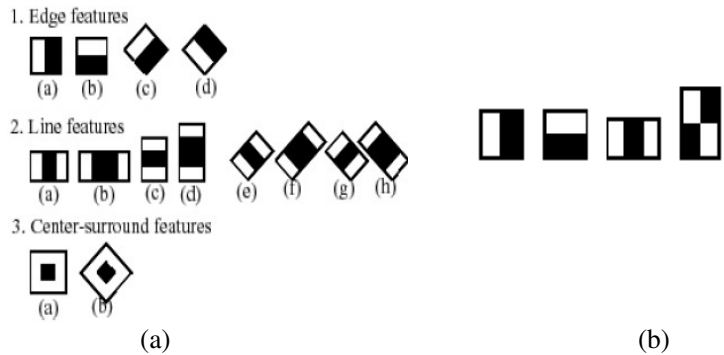

(a)                                                                                          (b)

Figure 2. Haar-like-features

Comparing with the raw pixel values, Haar-like features can reduce/increase the in-class/out-of-class variability, and thus making classification much easier. The rectangle Haar-like features can be computed rapidly using "integral image". Integral image at location of $x, y$ contains the sum of the pixel values above and left of $x, y$, inclusive:

$$P(x, y) = \sum_{x' \leq x, y' \leq y} i(x', y')$$

$$P_1 = A, P_2 = A + B, P_3 = A + C, P_4 = A + B + C + D$$

$$P_1 + P_4 - P_2 - P_3 = A + A + B + C + D - A - B - A - C = D$$

The sum of pixel values within "D":

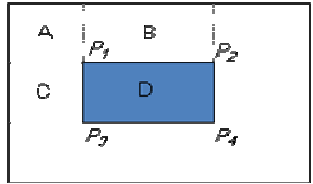

Figure 3. Haar-like features.

Using this Haar-like features the face detection cascade can be designed as in the figure 4, below. In this Haar cascade classifier an image is classified as a human face if it passes all the conditions, {f₁, f₂…, fₙ}. If at any stage any of one or more conditions is false then the image does not contain the human face.

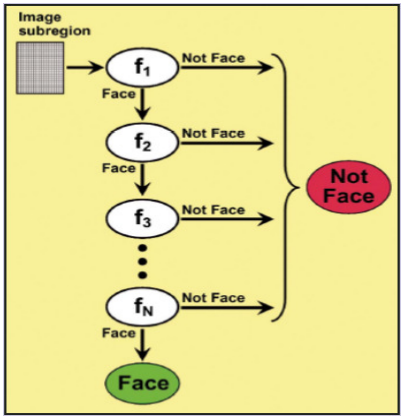

Figure 4. The cascade classifier classified face and non-face.

## 4.2. Face Recognition

### 4.2.1. PCA and Fast PCA (Principal Component Analysis)

Face recognition is one of the nonintrusive biometric techniques commonly used for verification and authentication. Local and global features [26] based extraction techniques are available for face recognition. Global features extraction technique calculates co-variance matrix of inter-images [27] whereas auto-correlation matrix is computed in local features technique. PCA is a mathematical procedure that transforms a number of possibly correlated variables into a smaller number of uncorrelated variables called principal components. PCA can be expressed in general as "a linear transformation of the image vector to the projection feature vector" as given below:

$Y = W^T X$, where, W is the transformation matrix having dimension *K x 1*, *Y* is the *K x N* feature vector matrix and *X* is the higher dimension face vector obtained by representing all the face images into a single vector

$$X = \{x_1, x_2, x_3, \ldots, x_N\}$$

Where, each $x_i$ is a face vector of dimension "$n$" obtained from the $M \times N$ dimension face image [28].

Table 1. Fast PCA algorithm for computing leading eigenvectors[22]

1. Choose , the number of principal axes or eigenvectors required to estimate. Compute covariance $\Sigma_x$ and set $\leftarrow 1$
2. Initialize eigenvector $\varphi$ of size $\times 1$ e.g. randomly
3. Update $\varphi$ as $\varphi \leftarrow \Sigma_x \varphi$
4. Do the Gram-Schmidt orthogonalization process

$$\varphi_p \leftarrow \varphi_p - \sum_{j=1}^{p-1} (\varphi_p^{\mathsf{T}} \varphi_j) \varphi_j$$

5. Normalize $\varphi$ by dividing it by its norm: $\varphi \leftarrow \varphi / \|\varphi\|$
6. If $\varphi$ has not converged, go back to step 3
7. Increment counter $\leftarrow +1$ and go to step 2 until equals

### 4.2.1 LDA

Linear Discriminant Analysis (LDA) finds the vectors in the underlying space that best discriminate among classes. For all samples of all classes the between-class scatter matrix $S_T$ and the within-class scatter matrix $S_W$ are defined. The goal is to maximize $S_T$ while minimizing $S_W$, in other words, maximize the ratio det$|S_T|$/det$|S_W|$ . This ratio is maximized when the column vectors of the projection matrix are the eigenvectors of ($S_W$^-1 $\times S_T$). The scatter matrices are defined as:

$S_T = \sum C\ N_i\ (ClasAvg_i$ - AvgFace $)\ (ClasAvg_i$  - AvgFace $)^T$

SW $= \sum C \sum X_i\ (x_k - SClasAvg_i\ )\ (x_k$  - $ClasAvg_i\ )^T$

where, C is the number of distinct classes, N is the number of images for each classes i, $ClasAvg_i$ is the average face image of face in class i, $X_i$ represents the face images that are in the class i, AvgFace is the average face image for all images in the database.

The algorithm for the LDA is given as below:

1. Represent the faces in the database in terms of the vector X.
2. Compute the average face AvgFace and subtract the AvgFace from the vector X.
3. Classify the images based on the number of unique subjects involved. So the number of classes, C, will be the number of subjects who have been imaged.
4. Compute the scatter matrix.
5. Use PCA to reduce the dimension of the feature space to N – C. Let the eigenvectors obtained be $W_{PCA}$.
6. Project the scatter matrices onto this basis to obtain non-singular scatter matrices $S_{WN}$ and $S_{BN}$.
7. Compute the generalized eigenvectors of the non-singular scatter matrices $S_{WN}$ and $S_{BN}$ so as to satisfy the equation $S_B * W_{LDA} = S_W * W_{LDA} * D$, where D is the eigenvalue. Retain only the C-1 eigenvectors corresponding to the C-1 largest eigenvalues. This gives the basis vector $W_{LDA}$.
8. Then the image vector X is projected onto this basis vector and the weights of the image are computed.

The advantages of using PCA along with LDA are as below:

1. Low memory required.

2. Low computational complexity.
3. Better recognition accuracy
4. Less execution time
5. Updating the inverse of the within class scatter matrix without calculating its inverse.

# 5. IMPLEMENTATION DETAILS

The overall system is divided into two basic modules: face detection and face recognition. There is a third module which uses the information of the above two modules. It is automated attendance system. This system is completely designed for the automated attendance system for the lab students, such as research scholars, research associates, M.Tech. Students, etc., to keep their practical records.

## 5.1 Face Detection

The Face Detection contains the following files:

**A. opencv1.sln**: This is a solution file which calls all other files. This .sln is created whenever we create a web application or any application in MS Visual Studio .Net. This file provides the editing facility in the code.
**B. prog1.cpp**: It is the main program file in the face detector module. It detects the face and crops the face image and saves in the current folder in which it is.
**C. haarcascade_frontalface_alt_tree.xml**: It is a cascade file in XML used to obtain Haar cascade for the frontal face in the image. It is used in the OpenCV library.
**D. StudentAttendence.xls**: It records the attendance of the detected face according to the system time in excel sheet.
**E. StudentAttendence.doc**: It is same as the above file; the only difference is that it saves the records in document format which can be easily printed for the detailed information.

## 5.2. Face Recognition

The Face recognition contains the following files:

A. **example.m**: It is the first page to be shown to the user. It calls the other files in this module. It takes input training dataset and also inputs the test dataset.
**B. CreateDatabase.m**: This module is in Matlab used to create database for the face images in the training dataset in a sequence of increasing numbers as the face images in the dataset are in number format.
**C. EigenfaceCore.m:** This module in the face recognition stage calculates the eigenface value using PCA and then applying the LDA algorithm on the result of PCA.
**D. facerec.m:** This creates graphical interface in Matlab for training and testing the database.
**E. Recognition.m:** This function compares two faces by projecting the images into face space and % measuring the Euclidean distance between them.
**F. facerec.exe:** This is the executable file created to linke the Matlab files with MS VS .NET 2008. It works in same way as the Matlab files does.

## 5.3. Automated Attendance

The attendance of each individuals entering in to the laboratory and going out from the laboratory is being recorded and an excel sheet is maintained. This excel sheet have various attributes, such as identified person's name, person's enrollment number, date of detection, time of detection, and detection and recognition time in milliseconds, which is useful for marking the attendance and deciding that the person should get full stipend or not. This module is made in MS VS .NET 2008.

# 6. RESULTS AND DISCUSSIONS

The system proposed is a real-time system. It takes input image through a web camera continuously till the system is shutdown. The captured image are then cropped by the Face Detection module and saves only the facial information in JPEG format of 100 x 100 matrix size. This is a colored image matrix having three layers. The layers are for red, green, and blue color in the image. The images are saved in a sequence of their occurrence time. That is, the face which is detected first is saved first in the database and the next is saved at the next place in database. The name of the face image is simply the numbers with extension .jpg. These numbers are the sequence number generated at the time of capturing. There are two factors for having file name as the number name. First is that it clearly indicates the sequence of the person they have come in-front of the camera. And the second factor is, at the time of training the system sequentially takes the training dataset of face images. It's very easy to create database of egienface using this method as any for loop is capable to increase the sequence number till the end of file. While if the file name is something, say text, then this would have been difficult to do. After creating the database the system is trained itself by calculating the face space. This is done by using the principal component analysis algorithm followed by linear disciminant analysis algorithm. These two algorithms are explained above. They reduce the dimension of the face space. These face space keeps on changing after each modification made to the TRAININGDATABASE. The image which is detected by the web camera are saved in another file/folder called TESTDATABASE, they are also in number.jpg format, e.g. 1.jpg, 194.jpg. number.jpg format, e.g. 1.jpg, 194.jpg.

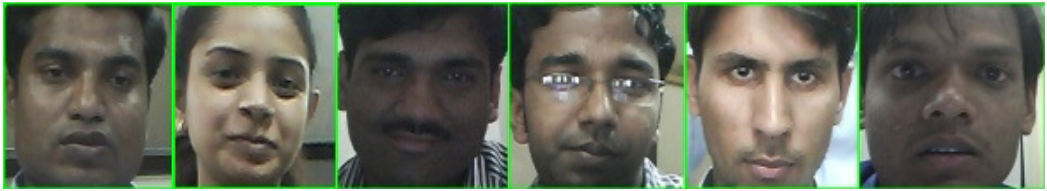

Figure 5. Images used in the system

In the database there are 500 samples, 5 images of each student in different position with different emotions. The face image used in this system is of 100 x 100 each. The images in the TESTDATABASE are used to test the system accuracy and to recognize the face from our database. Face recognition rate totally depend upon the database and the size of the image used. Also dimension of the image determines the accuracy rate of the system. In this paper, we studied and analyzed the facial features and extraction using fast PCA and LDA. Here, the comparison between PCA and LDA clearly show this.

*PCA < LDA:*
- **The training data set is large.**
- The number of training class is sufficient (using gallery).
- The number of feature is large (dimension).

*PCA > LDAL:*
- **The training data set is small.**
- The number of training class is not sufficient (using gallery).
- The number of feature is small (dimension).

The various output the proposed system are shown below one by one. In figure 6, the initial window that appliers will look like. In figure 7, the face detection is shown and in figure 8, the face recognition module is shown.

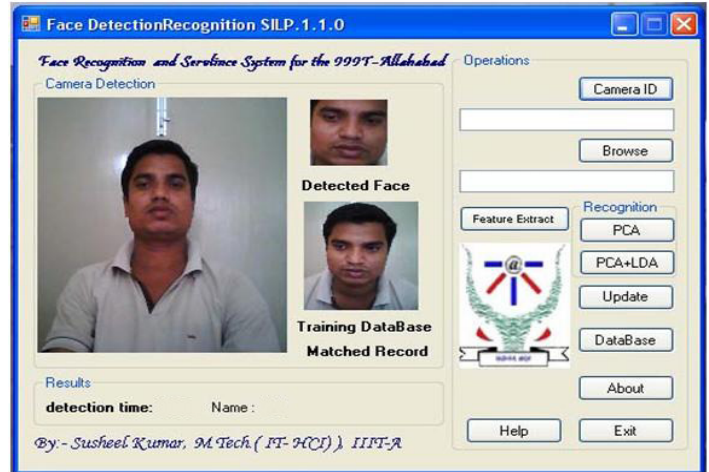

Figure 6. Initial screen of the system Application GUI.

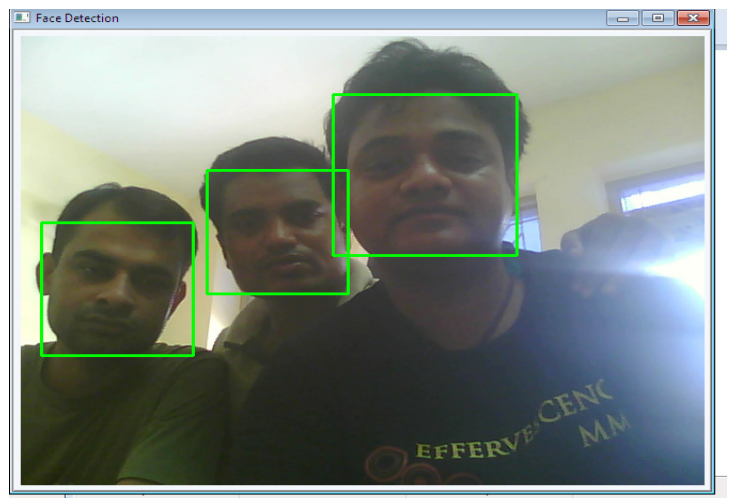

Figure 7. Face detection module.

| | A | B | C | D | E | F | G | H | I | J | K | L | M | N |
|---|---|---|---|---|---|---|---|---|---|---|---|---|---|---|
| 1 | | | | | | | | | | | | | | |
| 2 | | Student Online Attendence Record System | | | | | | | | | | | | |
| 3 | | | | | | | | | | | | | | |
| 4 | 1 | 21.jpg | Date and Time: | 4/16/2010 12:51 | Detection Time: | -107.263msec | | | | | | | | |
| 5 | 2 | 51.jpg | Date and Time: | 4/16/2010 12:51 | Detection Time: | 220.128msec | | | | | | | | |
| 6 | 3 | 81.jpg | Date and Time: | 4/16/2010 12:51 | Detection Time: | 160.898msec | | | | | | | | |
| 7 | 4 | 111.jpg | Date and Time: | 4/16/2010 12:51 | Detection Time: | 263.538msec | | | | | | | | |
| 8 | 5 | 141.jpg | Date and Time: | 4/16/2010 12:51 | Detection Time: | 588.961msec | | | | | | | | |
| 9 | 6 | 171.jpg | Date and Time: | 4/16/2010 12:51 | Detection Time: | -144.407msec | | | | | | | | |
| 10 | 7 | 201.jpg | Date and Time: | 4/16/2010 12:51 | Detection Time: | -67.0138msec | | | | | | | | |
| 11 | 8 | 231.jpg | Date and Time: | 4/16/2010 12:51 | Detection Time: | 148.195msec | | | | | | | | |
| 12 | 9 | 261.jpg | Date and Time: | 4/16/2010 12:51 | Detection Time: | -68.1643msec | | | | | | | | |
| 13 | 10 | 291.jpg | Date and Time: | 4/16/2010 12:51 | Detection Time: | -128.643msec | | | | | | | | |
| 14 | 11 | 321.jpg | Date and Time: | 4/16/2010 12:51 | Detection Time: | 571.319msec | | | | | | | | |
| 15 | | | | | | | | | | | | | | |
| 16 | | | | | | | | | | | | | | |
| 17 | | | | | | | | | | | | | | |

Table 5.2.3: real time face detection for attendance System file

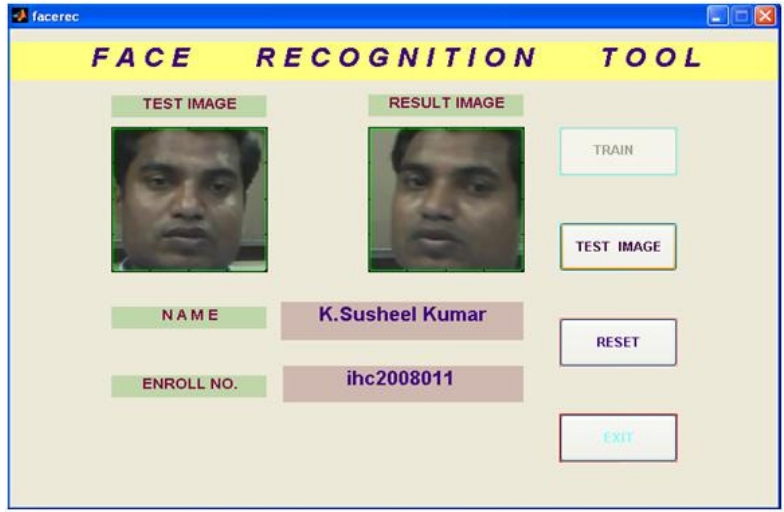

Figure 8. Face recognition module

## 7. CONCLUSION

The system has been tested on a wide variety of face images, with many emotions and many different angles other than frontal face image and to provide the security for person entry or out form the room or lab, whether the person is authorized or not. The system is giving very high accuracy. The system is capable of using multi cameras as the capturing device simultaneously and providing the detail of person of recognize whether is valid or not. If valid then record the time of the person and when person out from the room or lab then record the time of out .Thus, the system is good and efficient for general purpose like online attendance system into the class room or lab or any industries etc.

## ACKNOWLEDGEMENTS

The First I would like to thanks some great mind without whom this research would have been a distant reality. I am totally by the side of these people. I would like to say thanks to my parents who support to me carry out my research without any hindrance. My deepest thanks to great person, my mentor Prof. R.C.Tripathi and a big thanks to Mr. Shitala Prasad without whose ideas it was impossible last but not least to Mr. Vijay Bhaskar Semwal for excellent analysis of algorithm. I also extend my heartfelt thanks to my well wishers and unmentioned name.

## REFERENCES

[1]     Zhao, W., Chellappa, R., Phillips, P. J., Rosenfeld, A., 2003, Face recognition: A literature survey, ACM Computing Surveys (CSUR), V. 35, Issue 4, pp. 399-458

[2].    Elham Bagherian, Rahmita Wirza O.K. Rahmat, Facial feature extraction for face recognition: a review, Information Technology, 2008. ITSim 2008. International Symposium, Volume: 2, pp. 1-9

[3].    KIRBY, M. AND SIROVICH, L. 1990. "Application of the Karhunen-Loeve procedure for the characterization of human faces". IEEE Trans. Patt. Anal. Mach. Intell. 12

[4].    Elham Bagherian, Rahmita Wirza O.K. Rahmat, "Facial feature extraction for face recognition: a review," IEEE, 2008.

[5].    T. Kanade, "Picture processing by computer complex and recognition of human faces," technical report, Dept. Information Science, Kyoto Univ., 1973.

[6].     I.J. Cox, J. Ghosn, and P.N. Yianios, "Feature-Based face recognition using mixturedistance," Computer Vision and Pattern Recognition, 1996.

[7].     M. Lades, J.C. Vorbruggen, J. Buhmann, J.Lange, C. Von Der Malsburg, R.P. Wurtz, and M. Konen, "Distortion Invariant object recognition in the dynamic link architecture," IEEE Trans. Computers, vol. 42, pp. 300-311, 1993.

[8].     Shuicheng Yan, Huan Wang, Jianzhuang Liu, Xiaoou Tang, Huang, T.S. "Misalignment-Robust Face Recognition" Dept. of Electr. & Comput. Eng., Nat. Univ. of Singapore, IEEE Xplore , march 2010,vol 19, pages 1087 - 1096

[9].     L. Sirovich and M. Kirby, "Low-Dimensional procedure for the characterisation of human faces," J. Optical Soc. of Am., vol. 4, pp. 519- 524, 1987.

[10].    Xiaoyang Tan, Triggs. "Enhanced Local Texture Feature Sets for Face Recognition Under Difficult Lighting Conditions " Dept. of Comput. Sci. & Technol., Nanjing Univ. of Aeronaut. & Astronaut. Nanjing, China, IEEE computer science society, February 2010,vol 19,page 1635.

[11].    M. Kirby and L. Sirovich, "Application of the Karhunen- Loève procedure for the characterisation of human faces," IEEE Trans. Pattern Analysis and Machine Intelligence, vol. 12, pp. 831-835, Dec.1990.

[12].    Yin Zhang, Zhi-Hua Zhou, "Cost-Sensitive Face Recognition "Nat. Key Lab. for Novel Software Technol., Nanjing Univ., Nanjing, China IEEE, December 2009

[13].    L. Zhao and Y.H. Yang, "Theoretical analysis of illumination in pcabased vision systems," Pattern Recognition, vol. 32, pp. 547-564, 1999.

[14].    A. Pentland, B. Moghaddam, and T. Starner, "View-Based and modular eigenspaces for face recognition," Proc. IEEE CS Conf. Computer Vision and Pattern Recognition, pp. 84-91, 1994.

[15].    Yueming Wang, Jianzhuang Liu, Xiaoou Tang "Robust 3D Face Recognition by Local Shape Difference Boosting" Dept. of Inf. Eng., Chinese Univ. of Hong Kong, Hong Kong, China ,IEEE Xplore, January 2010

[16].    Belhumeur, V., Hespanda, J., Kiregeman, D., 1997, "Eigenfaces vs. fisherfaces: recognition using class specific liear projection", IEEE Trans. on PAMI, V. 19, pp. 711-720.

[17].    Roger (Ruo-gu) Zhang, Henry Chang, "A Literature Survey of Face Recognition And Reconstruction Techniques," December 12, 2005.

[18].    Y. Ryu and S. Oh, "Automatic extraction of eye and mouth fields from a face image using eigenfeatures and multiplayer perceptrons," Pattern Recognition, vol. 34, no. 12,pp. 2459–2466, 2001.

[19].    D. Cristinacce and T. Cootes, "Facial feature detection using adaboost with shape constraints," in Proc. 14th British Machine Vision Conference, Norwich, UK, Sep.2003, pp. 231–240.

[20].    L. Wiskott, J.M. Fellous, N. Kruger, and C. von der Malsburg, "Face recognition by elastic bunch graph matching," IEEE Trans. Pattern Analysis and Machine Intelligence,vol. 19, no. 7, pp. 775–779, 1997.

[21].    K. Toyama, R. Feris, J. Gemmell, and V. Kruger, "Hierarchical wavelet networks forfacial feature localization," in Proc. IEEE International Conference on Automatic Face and Gesture Recognition, Washington D.C., 2002, pp. 118–123.

[22].    T.F. Cootes, G.J. Edwards, and C.J. Taylor, "Active appearance models," IEEE Trans. Pattern Analysis and Machine Intelligence, vol. 23, no. 6, pp. 681–685, Jun. 2001.

[23].    J. Xiao, S. Baker, I. Matthews, and T. Kanade, "Real-time combined 2D+3D active appearance models," in Proc. IEEE Computer Society Conference on Computer Vision and Pattern Recognition, 2004, pp. 535–542.

[24].    Alok Sharma, Kuldip K. Paliwal, Fast principal component analysis using fixed-point algorithm, Journal Pattern Recognition Letters, Volume 28, Issue 10, 15 July 2007, Pages 1151-1155.

[25]. Zhiming Liu, Jian Yang, Chengjun Liu."Extracting Multiple Features in the CID Color Space for Face Recognition" Dept. of Comput. Sci., New Jersey Inst. of Technol., Newark, NJ, USA IEEE Xplore, April 2010,pages 2502 - 2509

[26]. Y. Freund and R.E. Schapire. A decision-theoretic generalization of on-line learning and an application to boosting. In Proceedings of the Second European Conference on Computational Learning Theory, pages 23–37. Springer-Verlag, 1995.

## Authors

**K Susheel Kumar** presently working as Assistance Professor in Ideal Institute of Technology, Ghaziabad, India. He is M.Tech form Indian Institute of Information Technology, Allahabad, his major research work Interest in Image Processing, computer sensor network & Pattern Recognition

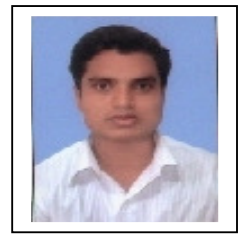

**Shitala Prasad** presently pursing his master degree in Information Technology from Indian Institute of Information Technology, Allahabad, India. He is B.Tech. Form IILM Greater Noida in Computer Science. His major research work interests in Image Processing and Gesture Recognition.

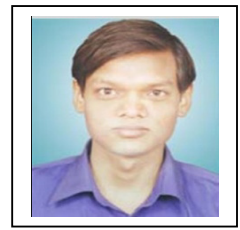

**Vijay Bhaskar Semwal** presently working with Newgen Software Technology, Noida, India as a Software Engineer in research Department. He is M.Tech. From Indian Institute Information Technology, Allahabad, his major research work interest in Wireless Sensor Network, Artificial Intelligence, Image Processing, Computer Network & Security and Design & Analysis of Algorithm.

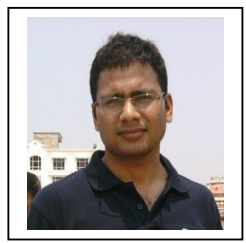

**Prof. R. C. Tripathi**,. Dean (R&D). Indian Institute of Information Technology Allahabad... VI-R&D in IT Experience his major research area intellectual property right (IPR), image processing, patten recognition

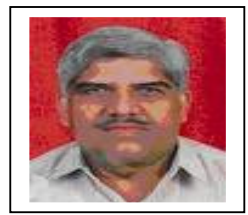